# Autonomous Formula Racecar: Overall System Design and Experimental Validation

Hanqing Tian, Jun Ni, Zirui Li and Jibin Hu*

## Abstract

This paper develops and summarizes the work of building the autonomous integrated system including perception system and vehicle dynamic controller for a formula student autonomous racecar. We propose a system framework combining X-by-wired modification, perception & motion planning and vehicle dynamic control as a template of FSAC racecar which can be easily replicated. A LIDAR-vision cooperating method of detecting traffic cone which is used as track mark is proposed. Detection algorithm of the racecar also implements a precise and high rate localization method which combines the GPS-INS data and LIDAR odometry. Besides, a track map including the location and color information of the cones is built simultaneously. Finally, the system and vehicle performance on a closed loop track is tested. This paper also briefly introduces the Formula Student Autonomous Competition (FSAC).

## Introduction

Formula Student Competition (FSC) is a famous competition in worldwide universities which encourages student to design, construct and compete with a Formula racecar. Organized and supported by Society of Automotive Engineering (SAE) and Institution of Mechanical Engineers (IMechE), FSC is well developed in U.S.A, Europe and China. Currently, FSC is combined by Combustion and Electric (FSEC) competition [1].

Recently, as autonomous ground vehicles (AGVs) has attracted focus in both academic and commercial applications [2, 3], combining AGV into FSC is essential for education purpose and the development of Formula Student Competition. In 2017, the first Formula Student Autonomous competition was successfully held in Hoffenheim, Germany. The author's team won the championship in the first Formula Student Autonomous China (FSAC), which is held with the help of SAE China. This paper summarized and reported our work on the FSAC, which mainly improves the performance of formulas dynamic control and path tracking behaviors under the rules of FSAC published by SAE China.

FSAC is a worldwide competition that combines several relative technologies of intelligent vehicle. The key technologies involve vehicle X-by-wire modification, system design of perception and motion planning according to the competition rules and dynamic control for autonomous formula. The three key technologies above are introduced and detailed in the next.

1) Different from FSC with a racing driver in the vehicle who can operate the racecar by mechanical system, FSAC need the vehicle controlled by electronic signal. Therefore, X-by-wire modification is the problem. [4-11] X-by-wired modification includes three parts: steering-by-wire, brake-by-wire and accelerate-by-wire. In [12], the authors complete the three parts' modification for an autonomous formula racecar based a traditional formula racecar.

2) The autonomous formula racecar is asked to race in specific environment for competition which is different from the traditional roads with lane line. According to the competition rule published by SAE China, the boundary consists of several cones (height: 45cm) with a lane width of 5 m approximately. [13] realizes the detection of traffic cones by the combination of two algorithms efficiently. Mapping and localization also need to be added in the framework of FSAC with the target of improving the length of the vehicle's path planning horizon [14].

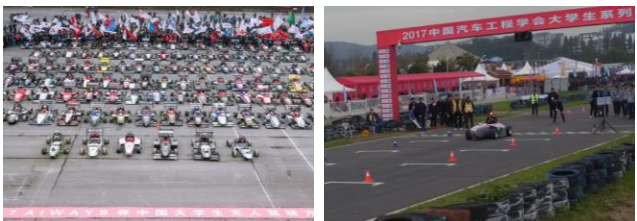

(a) Group Photo of All Teams      (b) Autonomous Race Test
**Figure 1.** Formula Student competition in China

National Key Lab of Vehicular Transmission, and School of Mechanical Engineering, Beijing Institute of Technology.

**Corresponding author:**
Jibin Hu, National Key Lab of Vehicular Transmission, and School of Mechanical Engineering, Beijing Institute of Technology, Haidian, Beijing 10008, China
Email: hujibin@bit.edu.cn

3) Autonomous formula racecar has its specific dynamic characteristic due to formulas competitive target, which is face with many differences in vehicle compared to the common autonomous vehicles for stabilization, safety and comfortable. Several vehicle control techniques are applied on formula racecar for path following [15-19]. [20-22] presented a controller combining Active Front Steer (AFS) and Direct Yaw Control (DYC) on formula racecar and improved the longitudinal and lateral tracking ability [23]. Krisada purposed a steering controller by understanding how a racecar driver controls a vehicle and using the center of percussion for an autonomous racecar. [24] developed a lane-keeping control system to assist even the best racecar driver in lane keeping.

However, most of the researches for the autonomous formula racecar focus on one or two part of the technologies above. But from the view of competition and system engineering, nearly research pay attention on the design of FSAC's system framework under the rules of competition. Individual vehicle modification, perception & motion planning system and dynamic control for path tracking cannot solve the problem in a racing competition. Therefore, with the target of racing and speeding and with the restriction of rules, a system framework combining the all three needed parts of FSAC becomes important which is not simply knocked together with three parts. Considering the competition of autonomous formula racecar, we develop and propose a FSAC system framework for autonomous formula racecar's design, perception and control which can provide a template which can be easily replicated.

The rest of this paper is organized as follows. Firstly, we introduce the vehicle Modification and software architecture. Then we introduce the algorithm including environment perception, planning and vehicle control in detail and presents the solution. Discussion of the experiment results are following. Conclusion and future work are presented in the last.

# Vehicle Platform and Software Architecture

The test vehicle platform named *Smart Shark* is an formula electric racing car (*BIT-FSE Silver Shark*) that our team have adapted and converted into an autonomous vehicle but retains the human-manipulated function. The vehicle keeps the fundamental features of a formula student racecar such as high handling stability. The distributed rear driven module is composed of double high power density wheel-side motors whose peak torque output can reach 80 Nm. The two independent motor controllers control the driven motors, respectively. Besides, the vehicle also has a high voltage tractive system with a peak power of 160 kW.

*A. Platform Modification*

The following works is needed to refit a FSE racecar to the autonomous racecar. Our team add servo system as steering and braking actuators because original manual driving racecar have no driving-by-wire system. We also reserve the manual driving facilities like mechanical steering linkage according to the rule which says the autonomous formula racecar should keep fully functional and suitable for human driving. The servo braking (SB) system and the redundant emergency braking system (EBS) are required in order to precisely control the brake force during deceleration and perform an emergency stop to ensure safety when failure of the system or emergency happen. The e-stop remote controller is required to trigger the braking system and shutdown HV supply from the tractive system remotely. It is manipulated by the ASR of a team. ASSI indicates the system state which is installed on the top of main loop. Computational devices and sensors are mounted in the positions according the rules. The modifications on the vehicle are shown in Fig. 2. Autonomous system master switch (ASMS) is designed to switch between driver and driverless mode by activating or shutdown self-driving system.

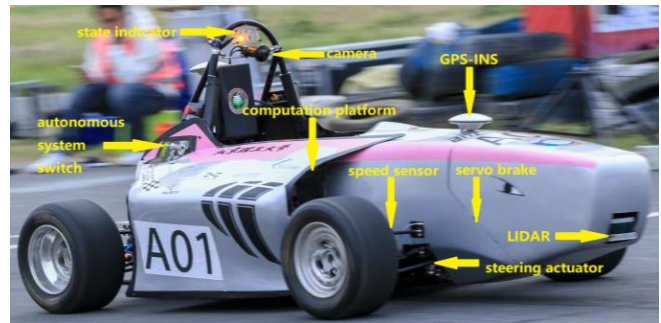

**Figure. 2**. Vehicle Modifications to an autonomous racecar

*1). Brake / Emergency Brake System*

According to the rules, we are supposed to design an emergency brake system. The structure, functional descriptions and control strategies are shown as follows. The structure of our EBS design is the linkage, which combines the servo and the brake pedal, shown as Figure 3 and 4.

One thing has to be mentioned is that when failure was detected i.e. after the EBS active and the oil press do not reach to the level that can perform a wheel lock, redundant actuator will switch on and active and works as the same states as the failure one.

To prevent ECU fail, the EBS back-up controller will always read a square wave(switch between 0 and 5V) in 50 Hz trough one I/O, and if it cannot read the level switch, which means that there is an error or a problem happening in ECU or software. And the EBS back-up controller will take over and control the EBS servo motor. The relationship between EBS back-up controller and ECU is independent. So we can confirm that the failure in ECU cannot influence the EBS's function.

In EBS, the actuators are powered by independent LV so that they can work when HV was shut down in emergency. Two redundant actuators are powered by to separate redundant LV supply. The control signal is sent by ECU CAN to the servos. As the control unit, our ECU is programmed to control the state of EBS. We also have a back-up controller for EBS as the redundancy. So, when the main ECU is failed, the back-up controller will take over and trigger the brake servo.

*2). Steering System*

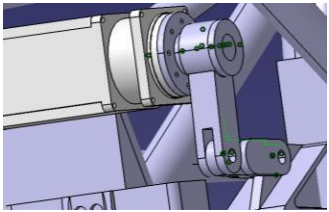
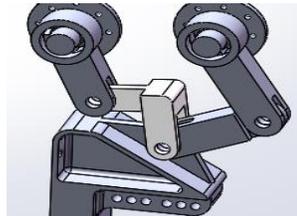

**Figure 3.** Brake actuator servo drives the linkage to pull the brake paddle.

**Figure 4.** Redundant brake system design to ensure safety

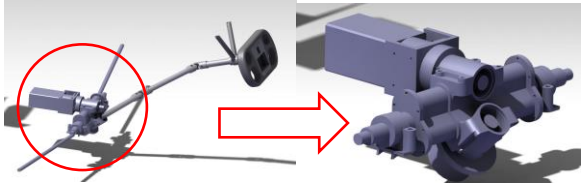

**Figure 5.** Steering system assembly

The steering system is designed to be able to work in manual steering mode and motor steering mode controlled by ECU. The steering rack was particularly designed into two sides and each side is designed with steering pinion. The torque working on the rack can come from steering wheel and motor, shown as Figure 5. Every time the LV active and vehicle starts, we need to choose a driving mode from the driver driving or automatic driving. ECU detects the ASMS switch signal to choose the driving mode.

### B. Software Architecture

The autonomous system is hierarchically separated to two levels which response for different levels of tasks. The overall architecture is shown in Fig. 6. The up-level system runs on a PC and ROS is installed as framware to manage all the data communications between nodes. The up-level software includes sensor driver node, detection node and planning and control node. The data of LIDAR, camera and GPS-INS on the vehicle is collected and synchronized according to time stamps by senor driver node. The main function of detection node is point cloud processing and image processing for cones detection and classification task. It also provides odometry for vehicle motion estimation and mapping. The planning and control node are for trajectory planning and motion control. We choose an industrial personal computer with high-perform computational resources and multiple hardware interfaces to meet the requirement. The low-level controller is for the actuator control task such as steering servo control. Up-level system gives control command such as steering angle and target acceleration to low level controller via serial port. In addition, the controller is also for the chassis safety control including ASSI, TS and EBS according to the competition safety rules. We choose a rapid-prototype vehicle control unit as the platform since its resources including 32-bit MCU with 264MHz router, multiple communication interface like CAN-BUS and serial port and multiple digital I/O can meet the requirement on computation and multi-interfaces.

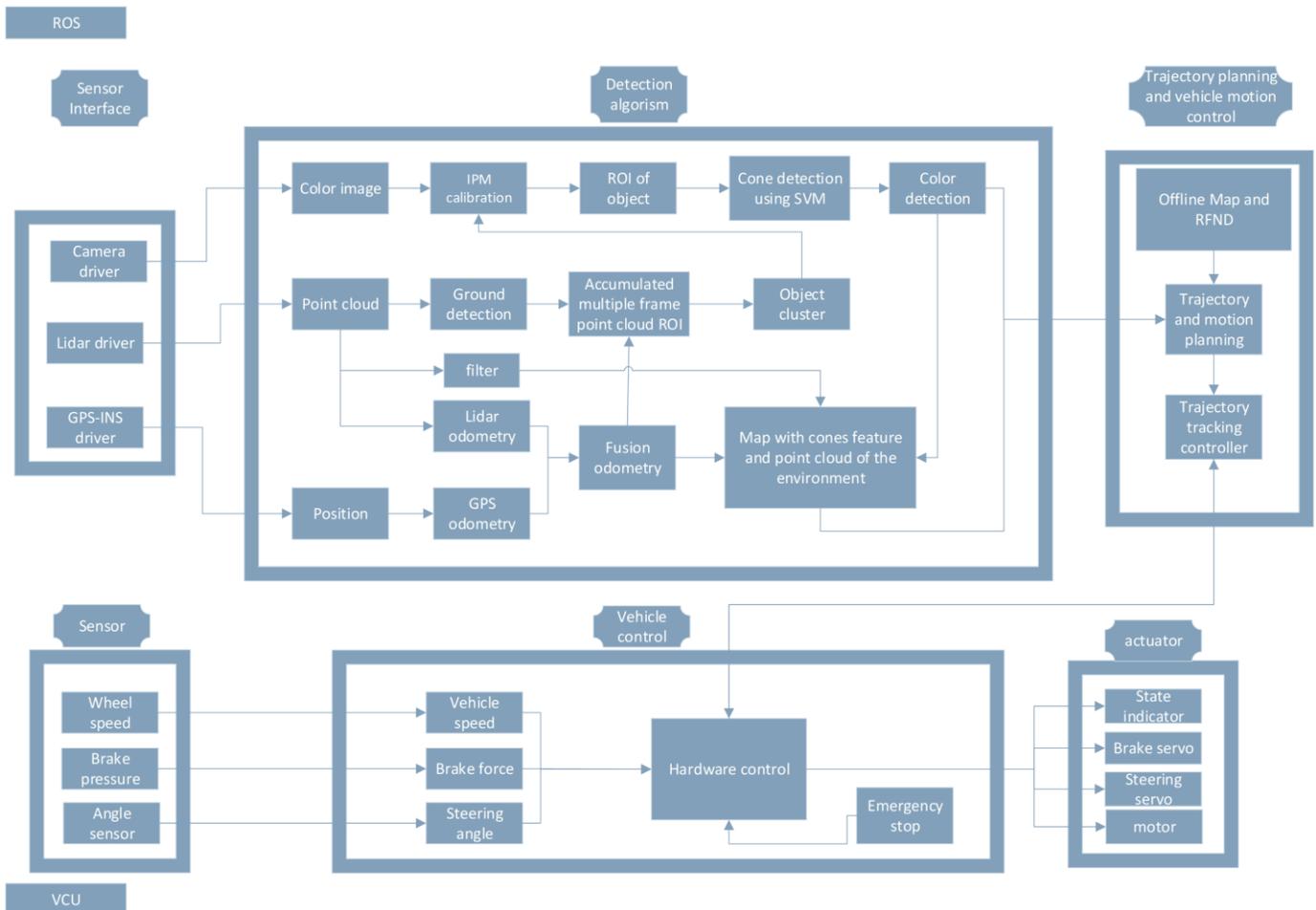

**Figure 6**. The detail flow diagram of the software pipeline.

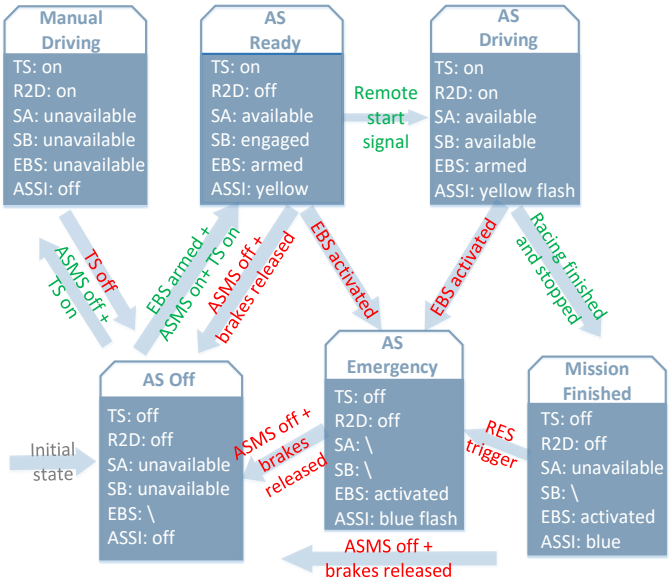

**Figure 7**. Autonomous system state machine

## C. Autonomous System State Machine and Function

Since autonomous system will implement perception and motion planning of an high speed unmanned vehicle by the algorithm based on the sensor implemented in the vehicle, software design need to ensure system safety, which is strictly regulated by the competition rules and need to be test in technical inspection by simulating some system failure modes like sensor and actuator failures. We design the autonomous system state machine according to the competition safety rule to define the subsystem working state at each time and the shifting condition between each state indicated by ASSI to ensure the vehicle behavior is explicit and controllable, in which way to ensure the safety of the track marshals no matter what emergency and error happen and lead the vehicle to a safe state. There are five states that will be implemented by autonomous system: "off", "ready", "autonomous driving", "Emergency (triggered)" and" finished". Fig.7 shows the function and working state of some subsystems of the racecar in each state and state transition conditions, respectively.

# Environment Perception

In the track, standard cones with 30cm high is used in FSAC as the main mark of the boundary of the track. The left boundary is marked by red cones while the right is marked by blue cones. Yellow cones are markers of start and finish line. Therefore, the position and color of cones need to be detected.

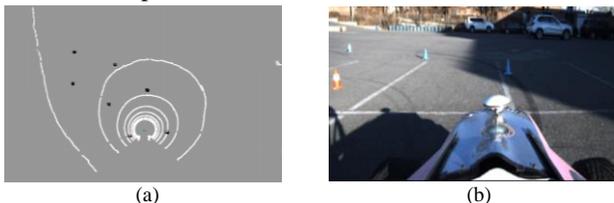

(a)              (b)

**Figure 8**. The white points are the inlier ground points and the dark points are the objects in (a) while (b) is the image from the camera.

Multiple sensors including LIDAR, camera and GPS-INS are cooperatively working in the mission.

### A. Laser Obstacle Detection

#### 1) Pointcloud Ground Segmentation

Since we only need to detect the cones ahead vehicle, ROI of the point cloud is chosen to be a box region in front of the car.

In order to extract object points, we firstly segment and remove the ground points. A straightforward strategy of ground and obstacle segmentation would be filtering pointcloud by a given threshold of vertical displacement. Indeed, this algorithm is widely used to segment large objects such as vehicles and pedestrians. However, during the situation with large acceleration, the pitch and roll error of the racecar body leads to the pose displacement between LIDAR and ground so the threshold cannot be set low enough in practice to segment small cone-sized objects without substantial numbers of false positives from ground points [25]. Actually, to avoid the false positives due to ground fluctuation during driving, we need to set a high threshold to filter the noise points.

A ground extraction strategy based on RANSAC algorithm is proposed to alleviate aforementioned problem. In our method, ground is modeled as a flat. Although a plane model is not able to represent terrains with large curvature and roughness, it is sufficiently effective to the track scenario for racecar since the environments is built of flat paving road which can be modeled as locally planar [26]. In our work, we build mathematical model of ground as a 3D plane. We also make the assumption that we only need to consider the height of the LIDAR from ground and the pose including the roll and pitch angle of the ground plane as parameter variable that need to be optimized. The model representation is written as follow:

$$y = ax + bz + h \qquad (1)$$

Where $h$ is a the proximately height of the LIDAR.

When every iteration begins, a hypothesis of plane model parameters is generated. Then all pointcloud data included in ROI region will be the observed data in hypothesis evaluation stage of RANSAC algorithm. There are the inlier datapoints whose distribution fit in the hypothesis model and the noise or outlier data which are the datapoints of cones outside plane. During every iteration, we define a criterion to distinguish whether a data point is an inlier data of the plane model under hypothesis parameters as:

$$L(p_i) = \begin{cases} 1 & if \ |error(p_i)| < \tau \\ 0 & else \end{cases} \qquad (3)$$

Error between points data and hypothesis flat model is calculated using the geometric distance,

$$error(p_i) = \frac{y - ax - bz - h}{\sqrt{1 + a^2 + b^2 + h^2}} \qquad (2)$$

Then the parameter including iterative time and threshold $\tau$ are set to iteratively search an optimized estimation of the model parameters by the optimization indicator which is to maximize the number of inlier points. Fig. 8 shows the result at an obvious steering condition which are chosen to validate if the method can distinguish obstacles effectively.

#### 2) Accumulating Multi-frame Pointcloud

After ground segmentation, the next stage should be the obstacle cluster process and position extraction. But as a matter of fact, the detection distance and the robustness can be limited by single frame pointcloud data only. Firstly, the laser scanner only has sixteen layers which leads to a large vertical gap between adjacent layers and low vertical resolution. It leaves a large non-detection space at long distance. Secondly, the small size obstacles with low height, for example cones in our scenario, will be hard to cluster since few datapoints are acquired and will be filtered as noise of the sensor. For purpose of increasing the effective detection range and detection robustness, multi-frame accumulation strategy is implemented which is analogous to the method of building HD map[27]. The accumulated pointcloud becomes dense enough for detection task.

The odometry which estimates the vehicle motion including heading and position relative to initial state is built to provide coordinate transfer for previous multi-frame pointclouds to current vehicle coordinate, which is the key step of the method at the first stage. In this paper, rigid body transformation matrix is calculated to represent the coordinate transfer. Wheel speed odometry get a good perform on wheel robotic system while it is limited on racecar due to large slip rate and tyre wear. GPS-INS can afford accurate positioning combined base station data but sometime is unstable. Some of the recent research about localization and mapping use visual odometry like mono and stereo camera or LIDAR odometry as the frontend[7]. In our work, LIDAR odometry and GPS-INS positioning are synchronized and coupled by complementary filter in order to get stable pose and motion estimation.

The LIDAR odometry method is separated into three steps inspired by LOAM which is a state-of-the-art odometry based on LIDAR sensor [28-30]. The problem definition of a LIDAR odometry is that: By given a set a sequence of pointcloud $\{\mathcal{P}_k\}$, $k \in Z^+$ arranged in chronological order, a method can utilize pointcloud registration algorithm to calculate the estimations of the three dimensional movement and pose transfer of the LIDAR at each time $t_k$. During the registration process, the feature points are extracted to minimize the calculation time and registration error. The method chooses the laser points in the convex edges and the flat plane as the features. To filter the features, an evaluation indicator of local smoothness is defined as follow:

$$c = \frac{1}{|S| \cdot \|X_{k,i}\|} \left\| \sum_{j \in S, j \neq i} (X_{k,i} - X_{k,j}) \right\| \quad (4)$$

where S represents the set of continuous laser scan data. Then edge feature points and plane feature points are selected by taking $c$ as the threshold.

The following step is the feature matching. Considering that the odometry algorithm operates in a frequency of 10Hz as same as the pointcloud data collection node, the displacement between adjacent time is small so that we can match each nearest matching line and surface defined by features in the pointcloud set $\bar{\mathcal{P}}_k$ to sharp edge features and flat plane features in $\bar{\mathcal{P}}_{k+1}$ as pair. Let **T** be the lidar pose transformation between $t_{k+1}$ and $t_k$. After obtaining the matching features pairs, we define the displacement error formula $d = f(\mathbf{T})$ between the matching lines and surfaces to the corresponding feature points. We estimate the transfer of LIDAR by minimizing the defined error between each feature point and its matching lines or surfaces. Levenberg-Marquardt method is implemented to solve this optimization problem as follow:

$$\mathbf{T} \leftarrow \mathbf{T} - \left(\mathbf{J}^T \mathbf{J} + \lambda \operatorname{diag}(\mathbf{J}^T \mathbf{J})\right)^{-1} \mathbf{J}^T d \quad (5)$$

Transferring and accumulating the pervious multiple frames of pointcloud to current LIDAR coordinate system $\widehat{L_k}$ at time $\hat{t}_k$ is the next stage as figure 9 shows. We define the quantity of superposition frame as $m$, each point in previous scan obtained at $t_k$ as $\mathcal{P}_k$, $k=1, 2 \ldots m-1$ and the points set acquired by multiple frame fusion as $\hat{\mathcal{P}}$. $L_k$ is the coordinate system of $\mathcal{P}_k$ and $X_{i,k}$ represents the coordinate of each point $i \in \mathcal{P}_k$ under $L_k$. Using the pose and motion estimation from odometry, the pose-motion transfer matrix $\mathbf{T_k}$ to the initial coordinate system can be solved as:

$$\mathbf{T_k} = \left[t_x, t_y, t_z, x, y, z, \omega\right]^T \quad (6)$$

Where $t_x$, $t_y$ and $t_z$ are translation displacement in direction of $x$, $y$, and $z$ axes while $x$, $y$, $z$, and $\omega$ are the quaternions to represent pose transfer. Points $\hat{X}$ in set $\hat{\mathcal{P}}$ is calculated as follow:

$$\widehat{X_{i,k}} = (\hat{\mathbf{R}} \mathbf{R_k}^{-1} (X_{i,k} - \mathbf{T_k}(1:3)) + \hat{\mathbf{T}}(1:3)) \quad (7)$$

Where $\hat{\mathbf{R}}$, $\mathbf{R_k}$ are the pose transfer matrix from initial coordinate system to LIDAR coordinate $\mathbf{L_k}$ and $\hat{\mathbf{L}}$, respectively. **R** is defined as follow when a pose and motion transfer matrix **T** is given:

$$\mathbf{R} = \begin{bmatrix} 1-2(y^2+z^2) & 2(xy-zw) & 2(xz+yw) \\ 2(xy+zw) & 1-2(x^2+z^2) & 2(yz-xw) \\ 2(xz-yw) & 2(yz+xw) & 1-2(x^2+y^2) \end{bmatrix} \quad (8)$$

*3) Object Detection Based on Euclidean Cluster*

The Euclidean clustering method is implemented to recognize object by grouping points on the same obstacle. Then algorithm calculates the envelope size and the centroid of each cluster points group. We filter the obstacle groups which are too large

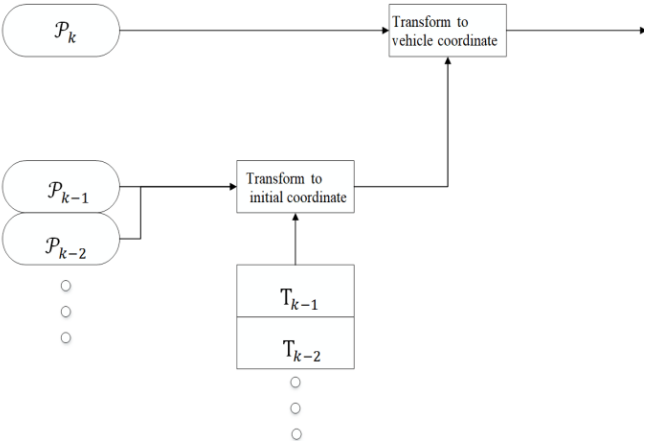

**Figure 9.** The flow chart of accumulating multi-frame. $\mathcal{P}_k$ is the pointcloud acquired at $t_k$. $\mathbf{T}_k$ is the pose and motion transfer matrix.

compared to the cone. The cone can be detected after clustering and size filtering.

However, there are other objects, such as a stack of tyres around that can be detected as cones. Therefore, visual information is used for verification discussed in the next section.

### B. LIDAR Assisted Visual Cone Classification

In this paper, a LIDAR assisted visual perception strategy is implemented to improve the computational speed to real-time level. Instead of ergodic processing every sliding windows on hole image, our method processes ROI regions pointed by the detection result from LIDAR only.

#### 1) Joint Calibration

This preparatory step gets parameters for projecting the image into the bird-view plane and projecting object position in vehicle coordinate system to image coordinate system via perspective transformation. The perspective matrix between two planes can be calculated by four disparate known point pairs:

$$\begin{bmatrix} x' & y' & w' \end{bmatrix} = \begin{bmatrix} x & y & w \end{bmatrix} \begin{bmatrix} \mathbf{M_1} & \mathbf{M_2} \\ \mathbf{M_3} & \alpha_{33} \end{bmatrix} \quad (9)$$

Where $x/w$ and $y/w$ represent the pixel position in the image coordinate. $x'/w'$ and $y'/w'$ represent the position of the cone in vehicle coordinate system. $\mathbf{M_1}, \mathbf{M_2}$ and $\mathbf{M_3}$ are defined as the linear transformation segment, perspective effect segment and translation segment, respectively. Figure 10 shows the bird-view after perspective transfer and the corresponding ROI bounding box delineated on image. It shows that the cluster result represented as dark points coincident with the bottom center of the cones on image plane.

#### 2) SVM Based Cones Detection

Gradient direction histogram is widely used to extract the object shape information as a rich feature set. In the track driving scenario, cones have distinct shape and edge feature. Therefore, we implement HOG features combined with SVM training classifier. The RGB image is converted to gray scale image and the gradient value in all directions at each pixel block is calculated. Finally, the length of the feature vector is generated from the gradient value, which can be used as the classification basis in the SVM classifier training. Linear kernel is selected as the kernel function. A dataset of the traffic cones used in the competition is built. It consists of 2500 positive samples and 9400 negative samples. Figure 11 shows the positive and negative samples from dataset. SVM algorithm is implemented by the library in OpenCV. Some positive samples which are the image of target and negative samples such as grass or curb are included in training dataset.

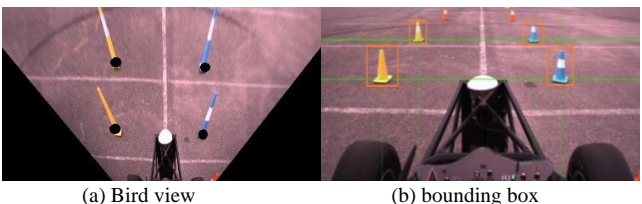

(a) Bird view          (b) bounding box
**Figure 10.** Joint calibration result by perspective transformation

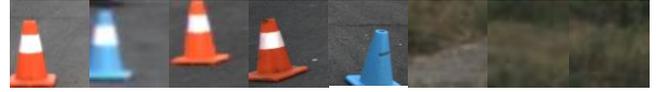

**Figure 11.** Some samples in training dataset.

#### 3) Cone color detection

Since SVM classifier can only determine if the ROI include cones, we still need to extract the color information. First, to reduce the influence of light on the camera image, the ROI is transferred to the HSV color space. In HSV space, H represents the hue. After that, the values of hue channel of each pixel are used as the input data of the k-means cluster. After taking k equal to two, image pixels are clustered into two groups based on color: pixels of the cones and pixels of the ground. We extract color from both pixel areas and determine the cone color.

### C. Mapping the Track and Detecting Close Loop

In the first run, the vehicle will operate the proposed real-time cones perception method. Since track cones are state obstacles, we can map the track during the first lap. For the vehicle position estimation, we use the odometry method which fused the information from GPS-INS and LIDAR using Kalman filter mentioned in section A. After fusion, the position estimation shows robustness under situation of bad GPS signal or LIDAR odometry failure.

With odometry and cones detection result, point cloud map of the track with cones marked as blue square can be built. shows in figure12.

In the end of the whole mapping process, the algorithm needs automatically detect the close loop when it returns to the start line, which is used to trigger the transition between detection driving mode and tracking driving mode and to count the number of driving lap. However, due to the mapping error and measuring drift of the sensor, a quantitative index is defined to determine the loop closure by comparing to the threshold determined by experiments. In our method, the quantitative coefficient $C$ of loop detection is shown as follow equation, which is consist of three kinds of error: the error of the map feature cones and detected cones currently, the error of the

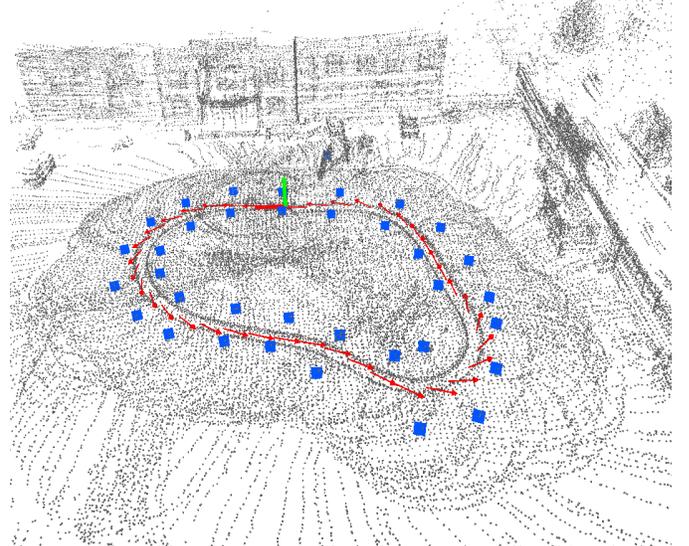

**Figure 12.** Cones are represented as blue box and red arrows show odometry data on the track map.

vehicle heading and vehicle position to the initial state:

$$C = W_c \cdot \sum_{k=0}^{j} \min_{i=0}^{j} \| X_{\text{detected}}^{(k)} - X_{\text{map}}^{(i)} \| + \quad (10)$$

$$W_h \cdot | H_{now} - H_{start} | + W_d \cdot \| L_{now} - L_{start} \|$$

Where $W_c$, $W_h$, $W_d$ are the weight parameter of the three parts of the error, respectively. $X$ is the coordinates of the cones. $H$ represent the heading of the vehicle while $L$ is the position.

## Motion Planning and Control

*A. Parametric Spline Curve based Trajectory Generation*

After the cones are detected and the track map is built, the car can navigate itself by tracking the track midline. The trajectory need pass through the middle point of each pair of cones with smooth curve for the vehicle to follow. The problem can be defined as: Given a set of points {P}, the method needs to define a synthetic curves S with n-order continuity while $\{P\} \subset S$.

In our work, parametric spline curve is implemented as the representation of the trajectory since it has been widely used to fit data points and the curve generated can interpolates Sequentially data points. Besides, compared to explicit nonparametric equation, parametric equation can represent a closed loop or multivalued sequential curve, like a closed loop course of the track. What's more, the trajectory can be discretized to a set of way point by sampling the variable and the first and second order derivative, namely, the heading slope and the curvature of the trajectory at each way point can be calculated. The general form of a parametric spline curve in 2-dimensional space is shown as:

$$\mathbf{P}(u) = [\mathbf{P}_x(u), \mathbf{P}_y(u)]_{1 \times 2}, u \in [0,1] \quad (11)$$

Here taking four distinct waypoints as the input as the fig.13 shows, the trajectory can be represented as cubic curve with parametric matrix **A**:

$$\mathbf{P}(u) = \begin{bmatrix} u^3 & u^2 & u & 1 \end{bmatrix}_{1 \times 4} \begin{bmatrix} a_{3x} & a_{3y} \\ a_{2x} & a_{2y} \\ a_{1x} & a_{1y} \\ a_{0x} & a_{0y} \end{bmatrix}_{4 \times 2} = \mathbf{U}_{1 \times 4} \mathbf{A}_{4 \times 2} \quad (12)$$

Given four points, we can solve the parametric matrix **A**:

$$\mathbf{A} = \begin{bmatrix} 0 & 0 & 0 & 1 \\ u_1^3 & u_1^2 & u_1 & 1 \\ u_2^3 & u_2^2 & u_2 & 1 \\ u_3^3 & u_2^2 & u_2 & 1 \end{bmatrix}^{-1} \begin{bmatrix} \mathbf{P}_0 \\ \mathbf{P}_1 \\ \mathbf{P}_2 \\ \mathbf{P}_4 \end{bmatrix} \quad (13)$$

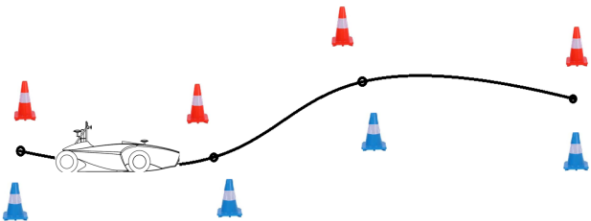

**Figure 13.** Four waypoints defined trajectory

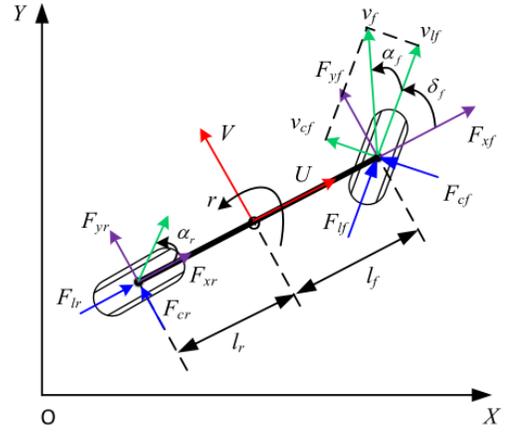

**Figure 14.** The 3-DoF vehicle model

So, the parametric curve can be written as:

$$\mathbf{P}(u) = \mathbf{UA} = \begin{bmatrix} u^3 & u^2 & u & 1 \end{bmatrix} \begin{bmatrix} 0 & 0 & 0 & 1 \\ u_1^3 & u_1^2 & u_1 & 1 \\ u_2^3 & u_2^2 & u_2 & 1 \\ u_3^3 & u_2^2 & u_2 & 1 \end{bmatrix}^{-1} \begin{bmatrix} \mathbf{P}_0 \\ \mathbf{P}_1 \\ \mathbf{P}_2 \\ \mathbf{P}_4 \end{bmatrix}$$

$$= \mathbf{UN}^s \mathbf{G}^s$$

$$= \mathbf{B}^s \mathbf{G}^s$$

$$= \begin{bmatrix} B_0^S(u), B_1^S(u), B_2^S(u), B_3^S(u) \end{bmatrix} \begin{bmatrix} \mathbf{P}_0 \\ \mathbf{P}_1 \\ \mathbf{P}_2 \\ \mathbf{P}_4 \end{bmatrix} \quad (14)$$

Where the $\mathbf{B}^s$ is the third-order basement matrix.

*B. Vehicle Control base on MPC*

This section describes the trajectory tracking controller for the unmanned formula car proposed in this paper. An MPC-based dynamics model is established, and the solver is used to solve the velocity and front wheel declination as the optimal control amount for the longitudinal and lateral directions. As shown in Fig. 14, the three-DOF dynamic model of the vehicle is built to establish the model predictive control model as a model basis for predicting the horizontal and vertical control of the unmanned formula car. The three degrees of freedom are the vehicle longitudinal speed, the lateral speed, and the yaw rate.

The state space equation built based on the three degree-of-freedom nonlinear vehicle dynamics model is as follows:

$$\dot{\xi} = f(\xi) + \mathbf{A}u_1 + \mathbf{B}u_2 \quad (15)$$

$$\xi = \begin{bmatrix} x \\ y \\ \psi \\ U \\ V \\ r \\ \delta_f \\ a_x \\ e_y \\ e_\psi \end{bmatrix} \quad f(\xi) = \begin{bmatrix} U\cos\psi - V\sin\psi \\ U\sin\psi - V\cos\psi \\ r \\ a_x \\ (F_{yf} + F_{yr})/M - Ur \\ (F_{yf}l_f - F_{yr}l_r)/I_{zz} \\ 0 \\ 0 \\ Ue_y + V \\ r \end{bmatrix} \quad \mathbf{A} = \begin{bmatrix} 0 & 0 \\ 0 & 0 \\ 0 & 0 \\ 0 & 0 \\ 0 & 0 \\ 0 & 0 \\ 1 & 0 \\ 0 & 1 \\ 0 & 0 \\ 0 & 0 \end{bmatrix} \quad \mathbf{B} = \begin{bmatrix} 0 & 0 \\ 0 & 0 \\ 0 & 0 \\ 0 & 0 \\ 0 & 0 \\ 0 & 0 \\ 0 & 0 \\ 0 & 0 \\ 0 & 0 \\ 0 & U \end{bmatrix}。$$

TABLE I
MPC PARAMETERS

| Parameters | Value |
|---|---|
| (x,y) | (x,y) Center of gravity in global coordinates |
| $\psi$ | Vehicle's heading angle |
| U | Longitudinal speed |
| V | Lateral speed |
| r | Yaw rate |
| $\delta_f$ | Steering angle |
| a | Longitudinal acceleration |
| $\varsigma_f$ | Steering rate |
| $J_x$ | Longitudinal jerk |
| M | Vehicle's mass |
| Izz | Moment of inertia |
| lf | Distances between the vehicles center of gravity and the front axle |
| $l_r$ | Distances between the vehicles center of gravity and the rear axle |
| $F_{yf}$ | Tire lateral forces generated at the front axle |
| $F_{yr}$ | Tire lateral forces generated at the rear axle |

The goal of trajectory tracking controller is to ensure the smoothness of the control and the safety of the vehicle within a predefined constraint. In order to achieve an optimal solution, the MPC-based trajectory tracking controller objective function is designed as follows:

$$J = \sum_k W_u(\delta_f^{(k)} - \delta_f^{(k-1)})^2 + \sum_k (W_{e\psi}(e_\psi^{(k)})^2 + W_{ey}(e_y^{(k)})^2) + \sum_k W_{sh} S_{sh}^{(k)} W_{sh} \quad (16)$$

The first term describes the steering stability of the vehicle, and the second term states that the third term of the vehicle's heading error and lateral position error is added as a relaxation factor to the objective function.

## Experiment

We recorded data in a track driving test. The track is a closed loop with a track width of 4 meters and a 5 meters distance between a pair of left and right cones shown in figure 15. Just like the track drive competition, the racecar needs to complete two complete cycles without any previous track data and automatically stop reaching the finish line during each test and shortest lap time is recorded.

On lap one, the racecar drives autonomously in low speed by realtime detection and records the track map by running odometry and mapping node simultaneously. Motion control is based on the proposed realtime trajectory generation algorithm that tracks the track midline shown in figure 16.

On lap two, racecar mainly relies on GPS-INS to localize itself and tracks waypoints generated on the previous lap. Proposed realtime detection and

TABLE II

| Parameters | MPC | Pure pursuit |
|---|---|---|
| Standard deviation of lateral acceleration(m/s²) | 0.1759 | 0.2338 |
| Lateral error of path following(m) | 0.2714 | 0.4520 |
| Average speed(m/s) | 2.9720 | 1.3677 |
| Average sideslip angle (Rad) | 0.0018 | -0.0120 |

trajectory planning run as the backup in order to handle the localization failure situations under bad GPS signal. The system can shift control strategy automatically to trajectory tracking mode when the closed loop trajectory detected. After the first lap, the system models the entire trajectory so that the vehicle runs steadily without being affected by the perception errors. What's more, the GPS-INS can provide high-frequency and accurate positioning information acquired without computational time delay during

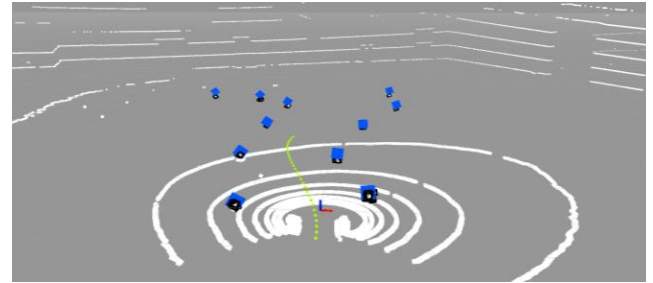
(a) cones cluster

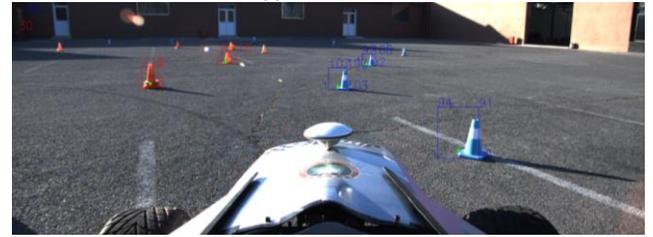
(b) Color detection
**Figure 16.** Mission on the first lap

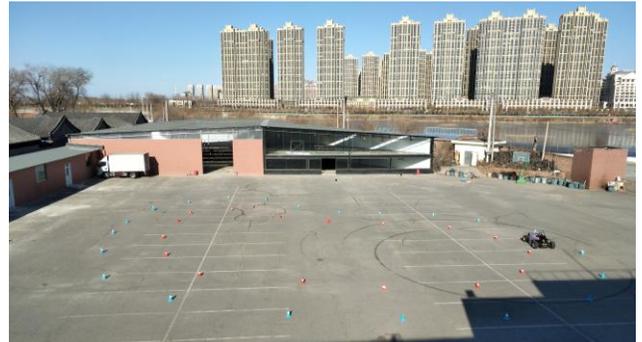
**Figure 15.** The test field environment and track

perception process. Therefore, second lap speed has a considerable improvement.

The authors' team won the FSAC champion in China. In the competition, the path tracking controller is design based on pure pursuit algorithm. The MPC-based controller applied on the formula racecar proposed in this paper is test compared to the pure pursuit algorithm. Steering angle, lateral acceleration, sideslip angle and lateral error during the path tracking behaviors are compared and analyzed in table II.

As shown in fig.17-20, the proposed autonomous formula racecar's path following controller has the same trend with pure pursuit-based path following controller in steering angle, lateral acceleration and lateral error. But the proposed controller can keep a higher speed than pure pursuit-based controller, which benefits to get a high score in the competition. For most of time, the proposed controller has a lower lateral path following error and a smoother lateral acceleration.

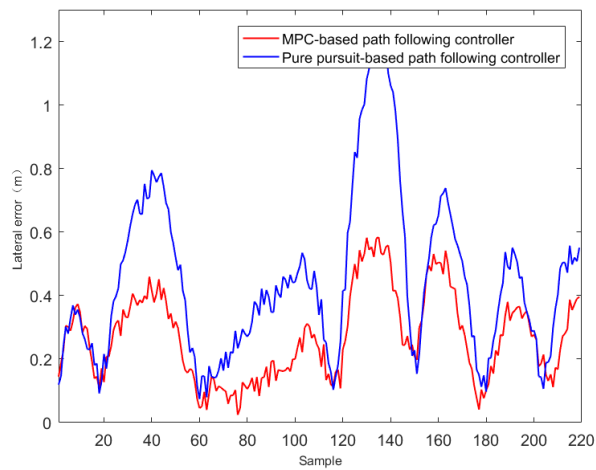

**Figure 18**. The tracking lateral error calculated from GPS-INS data. The MPC controller shows better performance on error response.

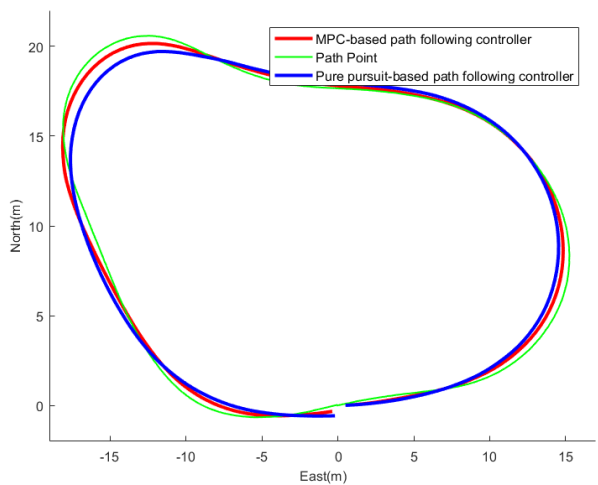

**Figure17**. The green curve is ground truth of the track middle line. The red and blue curve is tracking result by MPC and pure pursuit controller

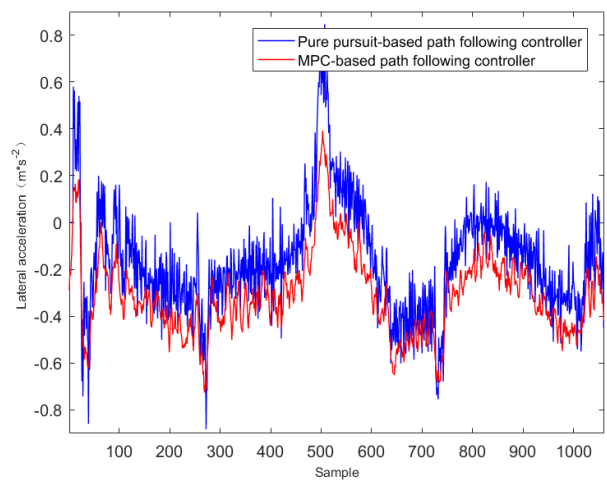

**Figure 20.** The lateral acceleration data.

## Conclusion

This paper reports our work for the development of the autonomous formula racecar. The racecar designed and developed by the authors' team is introduced. Overall system including perception, motion planning system and dynamic control for a FSAC racecar is discussed in detail. The X-by-wired system for steering, braking and accelerating is accurate enough to realize its own function with an expected performance that transcends the requirements of this project. It can be concluded that our proposed framework can handle the automatic driving mission under track environment of FSAC.

In the exist work, there are still several limitations. Firstly, in some environment short of feature points such as large playground, the LIDAR odometry will fail to registration. The similar situation will happen when there is occlusion issue of the LIDAR. Besides, Large moving obstacles in the track can also affect its registration due to feature movement and miss matching. In the future research, we are going to develop multi sensors fusion algorithm to enhance the odometry performance in featureless scenario. Secondly, the perception system shows noticeable time delay result from the limitation of the low algorithm efficiency and balance between the computational resource and the power consumption of the vehicle. We may try to use GPU acceleration technic to improve the efficiency under an acceptable power consuming.

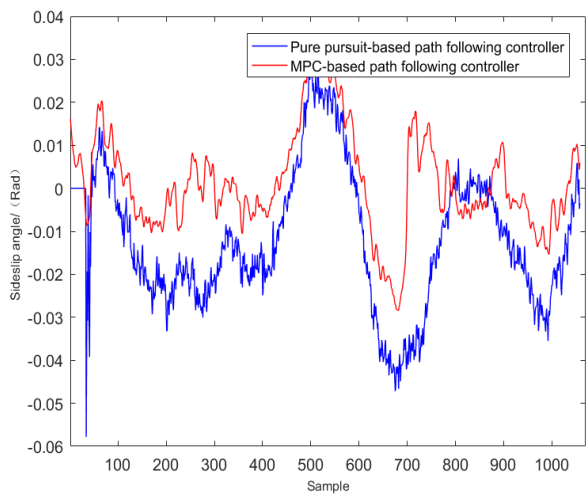

**Figure 19.** The Sideslip angle calculated from GPS-INS data.

# Appendix

*Abbreviations*

AS-Autonomous System
ASR-autonomous system responsible
ASMS-Autonomous System Master Switch
ASSI-Autonomous System Status Indicator
EBS-Emergency Brake System
ECU-Electronic Control Unit
FMEA-Failure Modes and Effects Analysis
LV-Low voltage system
HV-High voltage system
RES-Remote Emergency System
TS-Tractive system
R2D-ready-to-drive
SA-steering actuator
SB-servo brake